\PassOptionsToPackage{protrusion=true,expansion=false}{microtype}
\documentclass[11pt]{article}
\usepackage{acl}

\usepackage{graphicx}
\usepackage{amsmath}
\usepackage{amsfonts}
\usepackage{amssymb}
\usepackage[utf8]{inputenc}
\usepackage[T1]{fontenc}
\usepackage{lmodern}
\usepackage{hyperref}
\usepackage{url}
\usepackage{booktabs}
\usepackage{nicefrac}
\usepackage{microtype}
\usepackage{array}
\usepackage[table]{xcolor}
\usepackage{listings}
\usepackage{pifont}
\usepackage{multirow}
\usepackage{enumitem}
\hypersetup{pdftitle={Selective Intrinsic Self-Correction: From Revision Risk to Deployment Policy},pdfauthor={Tianzhu Zhang},pdfsubject={Intrinsic self-correction and selective revision},pdfkeywords={self-correction, language models, runtime gating}}

\newcommand{\stagezero}{Stage~0}
\newcommand{\stageone}{Stage~1}
\newcommand{\cw}{\ensuremath{\mathrm{C}\!\to\!\mathrm{W}}}
\newcommand{\wc}{\ensuremath{\mathrm{W}\!\to\!\mathrm{C}}}
\newcommand{\cc}{\ensuremath{\mathrm{C}\!\to\!\mathrm{C}}}
\newcommand{\ww}{\ensuremath{\mathrm{W}\!\to\!\mathrm{W}}}

\title{When Should LLMs Trust Their Own Revisions? A Risk-Aware Study of Intrinsic Self-Correction}
\author{Tianzhu Zhang\\
Nokia Bell Labs, Massy, France \\
Email: tianzhu.zhang@nokia-bell-labs.com
}

\begin{document}

\maketitle
\pagestyle{plain}
\thispagestyle{plain}

\begin{abstract}
Intrinsic self-correction asks a language model to revise its own answer without receiving new external evidence. A second pass can recover mistakes, but it can also overturn answers that were already correct. We study this trade-off across 29 open-weight LLMs on BoolQ, GSM8K, and Corr2Cause by tracking correctness transitions between initial and revised answers. Aggregate accuracy can conceal substantially different revision behavior: for example, Llama-3.1-8B improves by 25.5 percentage points on GSM8K while refinement changes 19.1\% of initially correct answers into wrong ones. A controlled BoolQ study further shows that refinement prompts shift the balance between recovery and harm. We then compare three runtime choices: keeping the initial answer, always accepting the revision, and selectively invoking revision using signals available after the initial response. The comparison identifies settings where learned gating is useful and others where a simpler unconditional policy performs better. These results suggest treating intrinsic self-correction as a revision policy rather than as a uniformly beneficial second pass, and evaluating it through both the corrections it recovers and the errors it introduces.
\end{abstract}

\section{Introduction}
Asking a language model to reconsider an answer also requires deciding whether the revision should replace the original. In intrinsic self-correction (ISC), the model receives no new external evidence, so an additional pass may repair an error or disturb an answer that was already correct. This matters whenever revision is invoked automatically, without a user checking both outputs. The system incurs the cost of another generation, and its policy determines which answer reaches the user. An additional answer is useful only if that policy retains enough corrections without introducing excessive errors or computation. The practical question is when revision is worth its cost and risk, and whether information available after the initial response is sufficient to make that decision.

Prior research establishes both the promise and the difficulty of this process. Language-feedback methods can improve outputs through iterative critique and refinement~\citep{madaan2023selfrefine,shinn2023reflexion}, while evaluations of frozen models show that unsupported reconsideration can reduce reasoning accuracy~\citep{huang2023large,kamoi2024can}. Mistake identification is itself a bottleneck. Correcting a located error does not imply being able to find it~\citep{tyen2024mistake}. \citet{yang2025confidence} decompose confidence and critique, making preservation and recovery distinct evaluation targets. Meanwhile, reinforcement learning can train models to verify and correct their own responses~\citep{kumar2025score,ma2025s2r}. These findings establish that correction depends on the available procedure and training. For a fixed procedure, the remaining runtime question concerns when to invoke it and which baseline should justify that decision.

Identifying an incorrect answer and predicting a useful revision are different tasks. An uncertain initial response may remain wrong after another pass, while a confident response may be correct and vulnerable to unnecessary revision. Variation between model--task settings can justify choosing different policies for those settings. The coexistence of recoveries and harmful flips within one setting creates an opportunity for instance-level selection. A useful gate also requires the consequences of revision to be predictable from features available before revision. Without that predictability, a gate can reject helpful changes as readily as harmful ones. Assessing the gate requires comparing its accuracy with both keeping every initial answer and revising every answer. Comparing only with the initial model would attribute improvements from the revision procedure itself to the gate.

We study these decisions across 29 open-weight models and three benchmarks covering reading comprehension, arithmetic reasoning, and causal inference: BoolQ, GSM8K, and Corr2Cause. The retained summary panel contains 82 model--task settings. We examine accuracy alongside the four possible correctness transitions and separately compare five refinement instructions on matched BoolQ initial answers. The recorded policy outcomes illustrate why both unconditional baselines matter. For Llama-3.1-8B on GSM8K, accuracy rises from 17.4\% to 43.0\% with unconditional revision and to 43.6\% with gating; most of that improvement comes from revision. For Qwen2.5-7B on BoolQ, unconditional revision instead reduces accuracy from 80.8\% to 72.3\%, whereas the recorded gate reaches 86.0\%. These examples distinguish improvements from revision from those from selection. The distinction depends on which changes the policy retains.

Our contribution is an empirical account of revision risk and the decision to incur it. First, we conduct a broad transition-level audit that distinguishes preserved correctness, persistent errors, harmful flips, and helpful recoveries across models, tasks, and prompts. Second, we connect this accounting to protocol validity. Record-level checks demonstrate how task-specific extraction, KEEP/CHANGE application, and reasoning-trace completion can alter measured outcomes. Third, we compare setting-level policy choice with pre-refinement instance gating, reporting cases where selection adds value, cases where an unconditional policy is preferable, and residual degradation. Together, these analyses examine invoking a fixed revision procedure using initial-answer information. They provide a basis for deciding whether revision risk is predictable enough to justify selective computation.

\section{Revision Protocol and Evaluation}
\label{sec:setup}
\subsection{Generation and scoring}
For each input $x$, the model first generates an initial response $y_0$. A revision input combines the original question, the previous answer, and a refinement instruction. The same model then generates a second response. The model receives no retrieval result, tool output, or oracle feedback. The revision payload depends on the protocol. Direct-final variants request a task-specific answer, verifier variants request KEEP or CHANGE, and reasoning variants generate a trace followed by a compact answer probe. Appendix~\ref{app:protocol} gives the prompt contracts.

Extraction follows the task and protocol variant. Binary tasks map explicit yes/no answers to labels; GSM8K normalizes and extracts a final integer. In a verifier-style run, the applied revised answer follows its KEEP/CHANGE rule, so a generated candidate need not become the final answer. We flag missing answers as invalid and score them incorrect in the recovered records. We retain these flags and include invalid outputs in the accuracy denominator. Comparing extracted answers with benchmark labels gives initial and revised correctness, from which we calculate transitions.

The panel comprises BoolQ validation (3,270 examples)~\citep{clark2019boolq}, GSM8K test (1,319)~\citep{cobbe2021training}, and Corr2Cause test (2,246)~\citep{jin2024can}. These sizes describe the benchmark splits; gate-summary denominators require their own matched records. The 82 retained summaries cover 29 open-weight checkpoints, with 28 BoolQ, 28 GSM8K, and 26 Corr2Cause settings. We include general instruction, code-oriented, math-oriented, and reasoning-tuned models. Complete initial, revised, gated, and selected-policy accuracies appear in Appendix~\ref{app:full-results}. Model coverage and the inference environment are detailed in Appendix~\ref{app:reproducibility}.

\subsection{Record-level protocol checks}
\label{sec:protocol-checks}
Two pairs of GSM8K generation files permit a direct validity check. Each pair contains 1,319 unique, matching IDs, consistent ground truth, and matching initial answers across stage files. For OLMo-3-7B, both stages contain 533 correct answers. The applied verifier rule accounts for this stability. The verifier produced 869 explicit KEEP decisions and 432 default KEEP decisions, covering 1,301 examples; the remaining 18 used direct application. Zero net change alone would not identify this mechanism.

For Phi-4-reasoning-plus, correctness falls from 1,220 to 1,216 answers, with 27 recoveries and 31 harmful flips. However, 650 initial and 546 revised traces reach the reasoning-token budget before natural completion. The measured transitions concern responses produced by budgeted traces followed by answer probes. They cannot be interpreted as changes between unrestricted completed reasoning traces.

\begin{table}[t]
\centering\small
\setlength{\tabcolsep}{3pt}
\begin{tabular}{lrr}
\toprule
GSM8K check & OLMo-3-7B & Phi-4-plus\\
\midrule
Matched items & 1,319 & 1,319\\
Invalid initial / revised & 18 / 13 & 2 / 4\\
Both answers parseable & 1,301 & 1,314\\
Correct initial / revised & 533 / 533 & 1,220 / 1,216\\
Helpful / harmful flips & 0 / 0 & 27 / 31\\
\bottomrule
\end{tabular}
\caption{Checks recomputed from paired records, retaining invalid outputs in the accuracy denominator. Phi-4-plus denotes Phi-4-reasoning-plus. These two checks do not certify the complete model panel.}
\label{tab:protocol-checks}
\end{table}

The OLMo run uses temperature 0, a 256-token generation budget, raw-prompt serialization, and conservative KEEP/CHANGE application. The Phi run uses temperature 0.6, its tokenizer's chat template, a 1,024-token reasoning budget, and an eight-token answer probe under the prefix \texttt{Final Answer:}. The recovered counts reproduce the corresponding initial and revised accuracies in the complete summary table. Other settings can use different protocols; these settings are not a shared decoding configuration for all 29 models.

\subsection{Correctness transitions}
Let $c_0(x) \in \{0,1\}$ denote whether the \stagezero{} answer is correct, and let $c_1(x) \in \{0,1\}$ denote whether the \stageone{} answer is correct. Each instance falls into one of four transition types:
\begin{align*}
\cc &: c_0=1, c_1=1 \quad \text{(correct preserved)}, \\
\cw &: c_0=1, c_1=0 \quad \text{(harmful flip)}, \\
\wc &: c_0=0, c_1=1 \quad \text{(helpful flip)}, \\
\ww &: c_0=0, c_1=0 \quad \text{(persistent error)}.
\end{align*}

We report standard accuracy metrics both before and after a round of refinement.
\[
\mathrm{Acc}_0 = \frac{1}{N}\sum_x c_0(x), \qquad
\mathrm{Acc}_1 = \frac{1}{N}\sum_x c_1(x)
\]
\[
\Delta = \mathrm{Acc}_1-\mathrm{Acc}_0
\]
Only two transitions change accuracy. A \wc{} transition adds one correct answer, while a \cw{} transition removes one, giving
\[
\Delta = \frac{\#\wc-\#\cw}{N}
\]
Accuracy change alone does not reveal the conditional risk of revision. We also report
\[
\mathrm{HarmRate} = \frac{\#\cw}{\#\cc+\#\cw}
\]
\[
\mathrm{RecoveryRate} = \frac{\#\wc}{\#\wc+\#\ww}
\]
HarmRate measures how often refinement damages an initially correct answer, while RecoveryRate measures how often it repairs an initially wrong answer. These conditional rates explain a useful base-rate effect:
\[
\begin{aligned}
\Delta
&= (1-\mathrm{Acc}_0)\,\mathrm{RecoveryRate} \\
&\quad - \mathrm{Acc}_0\,\mathrm{HarmRate}
\end{aligned}
\]
Thus, the same HarmRate and RecoveryRate can have different net effects depending on \(\mathrm{Acc}_0\). For an LLM with high initial accuracy, the initially correct pool is large, and the initially wrong pool is small; consequently, even a small HarmRate can cancel a much larger RecoveryRate. This base-rate effect is especially relevant for reasoning models that already answer most questions correctly.

\section{How Revision Changes Correctness}
\label{sec:characterization}
\subsection{Revision gains vary across settings}
Unconditional revision reduces accuracy in 20 of 28 BoolQ settings, 10 of 28 GSM8K settings, and 9 of 26 Corr2Cause settings in the retained summary panel. Revision also produces substantial gains. Llama-3.1-8B improves by 25.55 percentage points on GSM8K, and Qwen3-Coder-30B-A3B improves by 27.64 points on Corr2Cause. The GSM8K results contain both large improvements and losses.

\begin{figure*}[t]
\centering
\includegraphics[width=\textwidth]{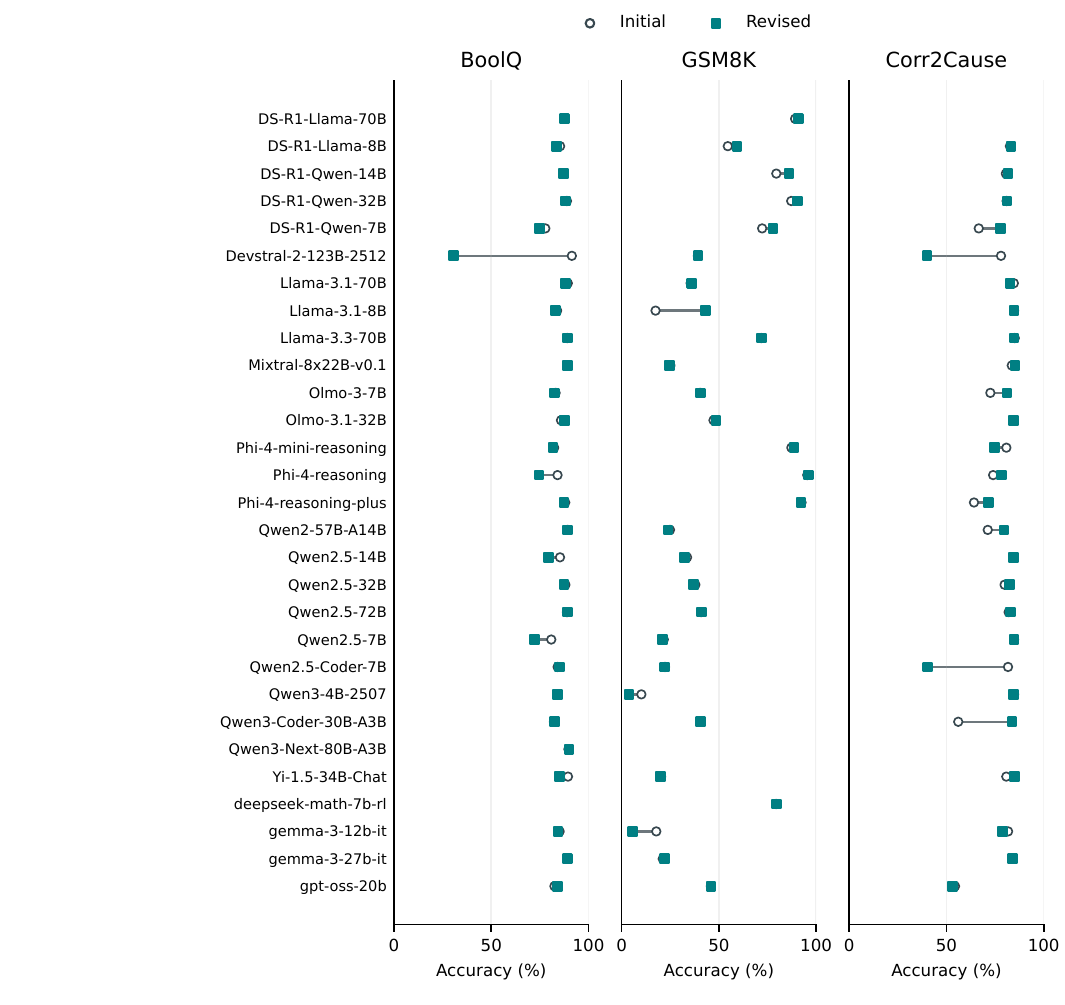}
\caption{Initial and revised accuracy from the same 82-row summary source used by the complete tables in Appendix~\ref{app:full-results}. Each segment joins one setting's two accuracies; blank positions indicate missing settings. DS-R1 abbreviates DeepSeek-R1-Distill, and instruction suffixes are shortened. The earlier diagnostic plots use different run summaries and are shown separately in Appendix~\ref{app:diagnostic-panels}.}
\label{fig:two-stage-isc}
\end{figure*}

Between-setting variation argues for comparing policies separately for each model--task--protocol combination. It does not by itself show that instance gating is useful. For example, a consistently effective revision procedure may justify always revising; selection adds value only if it improves the accuracy--cost trade-off within that setting.

\subsection{Net gains can conceal harmful revisions}
The largest net gain in Table~\ref{tab:transition-metrics} coexists with substantial harm to initially correct answers. In the Llama-3.1-8B GSM8K transition summary, 381 recoveries outweigh 44 harmful flips, giving a 25.5-point gain. Yet those flips overturn 19.1\% of the initially correct pool. Count-based 95\% bootstrap intervals are [22.9, 28.4] points for the gain and [14.2, 24.3]\% for conditional harm. DeepSeek-R1-Distill-Qwen-14B instead combines a 6.5-point gain [4.6, 8.4] with 4.2\% conditional harm [3.0, 5.5]. These intervals condition on the reported counts and do not include run-selection uncertainty.

\begin{table*}[t]
\centering
\scriptsize
\setlength{\tabcolsep}{4pt}
\begin{tabular}{@{}llrrrrrrrl@{}}
\toprule
& & \multicolumn{4}{c}{Transition counts} & \multicolumn{3}{c}{Rates / delta (\%)} & \\
\cmidrule(lr){3-6}\cmidrule(lr){7-9}
Model & Task & \cc & \cw & \wc & \ww & Net $\Delta$ & Harm & Rec. & Takeaway  \\
\midrule
Olmo-3.1-32B-Instruct 
& BoolQ 
& 2767 & 40 & 96 & 367 
& +1.7 & 1.4 & 20.7 
& low-harm gain \\

Llama-3.1-70B-Instruct 
& BoolQ 
& 2762 & 159 & 122 & 227 
& -1.1 & 5.4 & 35.0 
& net harmful \\

\addlinespace[1pt]
Llama-3.1-8B-Instruct 
& GSM8K 
& 186 & 44 & 381 & 708 
& +25.5 & 19.1 & 35.0 
& high-gain/high-harm \\

DeepSeek-R1-Distill-Qwen-14B 
& GSM8K 
& 1005 & 44 & 130 & 140 
& +6.5 & 4.2 & 48.1 
& gain/lower-harm \\

\addlinespace[1pt]
DeepSeek-R1-Distill-Qwen-7B 
& Corr2Cause 
& 1446 & 50 & 301 & 449 
& +11.2 & 3.3 & 40.1 
& broad recovery \\

Phi-4-reasoning-plus 
& Corr2Cause 
& 1392 & 49 & 213 & 592 
& +7.3 & 3.4 & 26.5 
& reasoning-model gain \\
\bottomrule
\end{tabular}
\caption{Representative transition-count summaries. Counts sum to the nominal task split sizes; Harm and Rec.\ are conditional rates defined in Section~\ref{sec:setup}. These summaries and the gate-evaluable accuracy panel have different provenance and are not treated as item-matched gate records.}
\label{tab:transition-metrics}
\end{table*}

A negative net change can also contain useful corrections. Llama-3.1-70B on BoolQ recovers 122 errors but overturns 159 correct answers. Its loss reflects the difference between those two populations, not a lack of correction capability. Whether a gate can retain the recoveries while blocking damage depends on information available before the second pass.

\subsection{The refinement instruction changes the trade-off}
\label{sec:prompt-sensitivity-results}
The matched prompt experiment is restricted to BoolQ. It evaluates 11 models under five refinement instructions, with initial answers held fixed. All five prompt families have negative mean accuracy changes, with the neutral prompt showing the smallest loss. More forceful instructions to switch answers increase harmful transitions more than helpful ones. The experiment shows that the available revision operation depends on its instruction, even when the base model and initial responses are unchanged.

\begin{figure}[t]
\centering
\includegraphics[width=\linewidth]{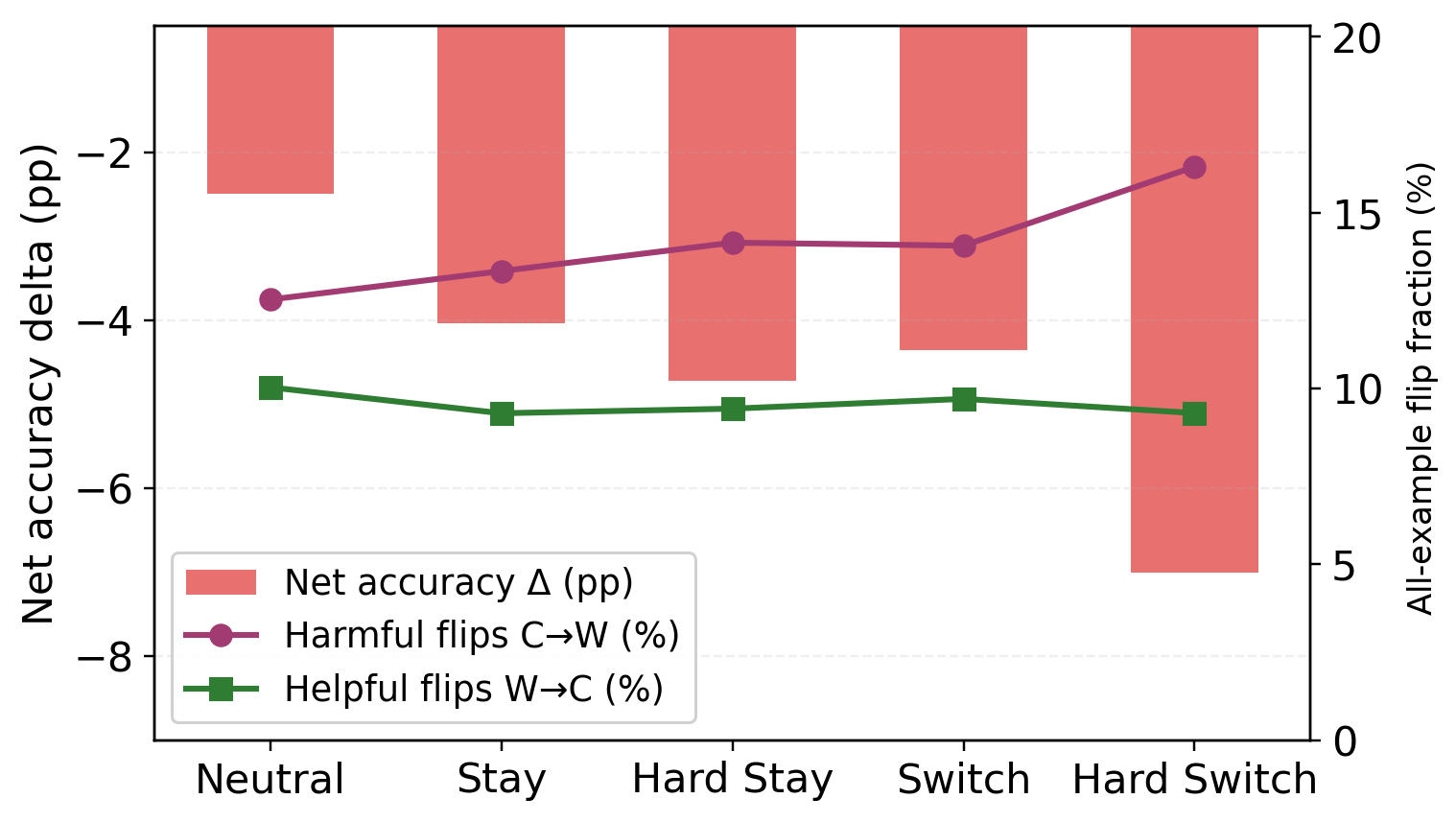}
\caption{BoolQ prompt comparison across 11 models. Bars show mean accuracy change. The harmful and helpful curves show joint transition fractions, $\#\cw/N$ and $\#\wc/N$, rather than conditional HarmRate and RecoveryRate. Their difference gives the accuracy change. No prompt-level uncertainty intervals are available in these plotted summaries.}
\label{fig:prompt-sensitivity}
\end{figure}

\begin{figure*}[t]
\centering
\includegraphics[width=.95\textwidth]{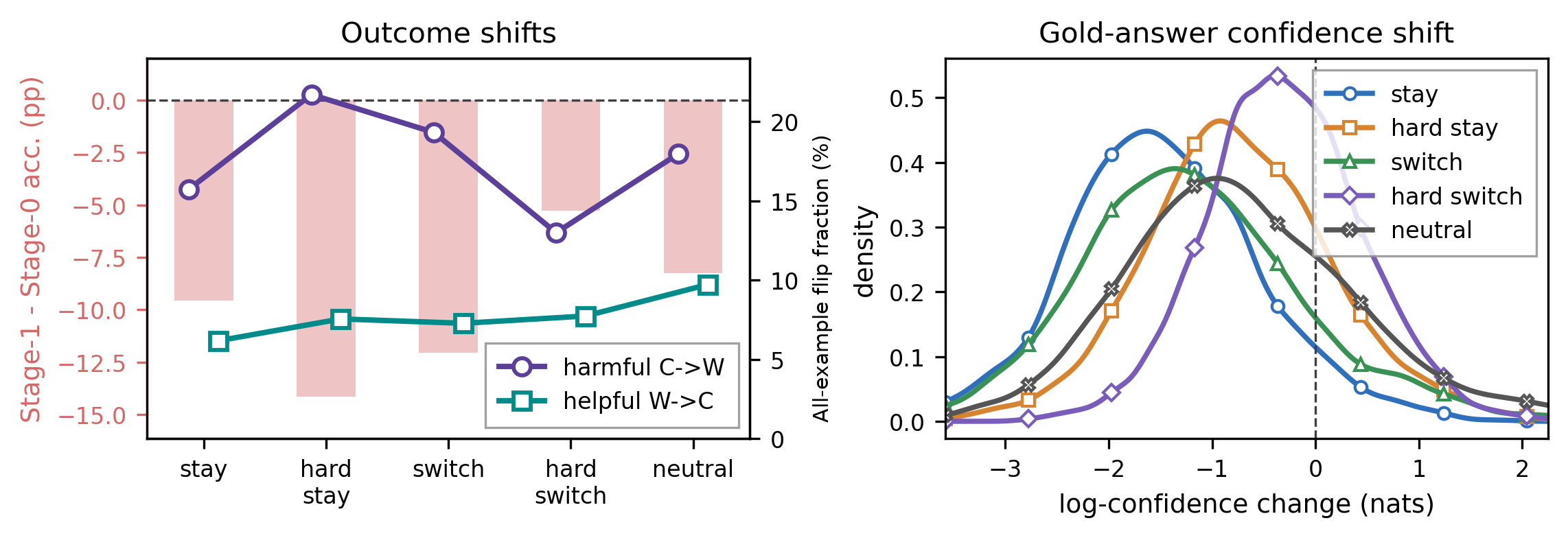}
\caption{Qwen2.5-7B on matched BoolQ initial answers. Left: accuracy change and joint harmful/helpful fractions of all evaluated examples. Right: the distribution of changes in diagnostic gold-answer log confidence; negative values indicate reduced confidence in the correct answer. Gold-answer confidence is an analysis diagnostic, not a deployable gate input.}
\label{fig:prompt-case-qwen25}
\end{figure*}

Qwen2.5-7B illustrates the effect within a single model (Figure~\ref{fig:prompt-case-qwen25}). Prompt wording shifts the accuracy change and the mix of revisions. The gold-answer confidence distributions also place substantial mass below zero, showing that reconsideration can reduce support for the correct label. These observations identify where revision changes correctness, but they do not show whether beneficial changes can be selected in advance. We next examine that decision using initial-answer features.

\section{Choosing When to Revise}
\label{sec:gating}
\subsection{Policies and runtime decisions}
A setting-level selector chooses one policy for a fixed model, task, and revision protocol: always keep \stagezero{}, always invoke \stageone{}, or use an instance-level gate. The gate $g(x)$ observes the initial response and its generation features before deciding whether to invoke the fixed revision procedure:
\[
y_g(x)=
\begin{cases}
y_1(x), & g(x)=1,\\
y_0(x), & g(x)=0.
\end{cases}
\]
Here $y_1$ is the applied revised answer defined by the protocol. A gate using only initial-answer information can save revision computation. A rule that first generates $y_1$ and then chooses an answer is an acceptance policy and cannot claim the same generation savings.

For a common evaluation population of size $N$, the gate's accuracy gain over the initial model is
\[
\begin{aligned}
\mathrm{Acc}_g-\mathrm{Acc}_0
&=\frac{A_{\wc}-A_{\cw}}{N},
\end{aligned}
\]
Its gain over unconditional revision is instead.
\[
\mathrm{Acc}_g-\mathrm{Acc}_1
=\frac{B_{\cw}-B_{\wc}}{N},
\]
where $A$ and $B$ count accepted and blocked transitions of the indicated type. The second identity distinguishes revision value from the incremental value of gating. A gate can improve on the initial model yet underperform unconditional revision if it blocks too many recoveries.

\subsection{Features, fitting, and result provenance}
The implementation extracts confidence summaries, parser and answer-format state, generation metadata, and agreement between parsed and scored candidate answers from \stagezero{}. Gold correctness defines training targets for helpful and harmful transitions, but is not a runtime input. We remove missing or nearly constant feature columns using the training data. Candidate classifiers include logistic regression, HistGradientBoosting, and LightGBM; Appendix~\ref{app:gate-details} describes the search.

The recovered version-2 trainer selects model configurations and thresholds using out-of-fold predictions on the training split. It refits the chosen predictor on that split and applies the frozen threshold to a separately supplied evaluation split. This train-CV/evaluation design supports independent testing without a third partition, provided example IDs are disjoint, and evaluation labels do not guide later decisions. The script reports ID overlap but does not automatically reject it. Earlier recovered trainers instead search thresholds on their supplied validation split.

The reported gate and setting-level selector outcomes remain exploratory. Saved run configurations and selected-policy artifacts have not been matched to every summary row, so the newer trainer cannot retrospectively establish their evaluation independence. The main figures, aggregate statistics, and complete policy tables are generated from one 82-row transcription. The separate accepted-transition summaries support algebraic checks, but do not supply matched gate IDs or denominators. Section~\ref{sec:limitations} discusses how these evidence gaps limit interpretation, and Appendix~\ref{app:gate-details} reports the numerical audit.

\subsection{Incremental value and failure modes}
\begin{table*}[t]\centering\fontsize{9}{11}\selectfont
\setlength{\tabcolsep}{3pt}
\begin{tabular}{llrrrrl}\toprule
Dataset & Model & Initial & Revised & Gate & Selected & Policy\\\midrule
\multicolumn{7}{l}{\textit{Gate exceeds both unconditional baselines}}\\
GSM8K & Llama-3.1-8B & 17.44 & 42.99 & \textbf{43.59} & \textbf{43.59} & gate \\
Corr2Cause & Qwen2-57B-A14B & 71.24 & 79.52 & \textbf{79.70} & \textbf{79.70} & gate \\
BoolQ & Qwen2.5-7B & 80.83 & 72.32 & \textbf{86.02} & \textbf{86.02} & gate \\
BoolQ & gpt-oss-20b & 82.45 & 84.13 & \textbf{85.08} & \textbf{85.08} & gate \\
\midrule
\multicolumn{7}{l}{\textit{Recorded selector chooses an unconditional policy}}\\
Corr2Cause & Qwen3-Coder-30B-A3B & 56.06 & \textbf{83.70} & 56.06 & \textbf{83.70} & S1 \\
Corr2Cause & DS-R1-Qwen-7B & 66.61 & \textbf{77.78} & 71.28 & \textbf{77.78} & S1 \\
Corr2Cause & Yi-1.5-34B-Chat & 80.77 & \textbf{84.86} & 83.48 & \textbf{84.86} & S1 \\
Corr2Cause & Phi-4-reasoning & 74.00 & \textbf{78.32} & --- & \textbf{78.32} & S1 \\
GSM8K & DS-R1-Qwen-32B & 87.19 & \textbf{90.30} & 89.84 & \textbf{90.30} & S1 \\
GSM8K & Qwen2.5-7B & \textbf{21.68} & 20.92 & 20.92 & \textbf{21.68} & S0 \\
\midrule
\multicolumn{7}{l}{\textit{Residual degradation: selected accuracy below initial accuracy}}\\
BoolQ & DS-R1-Llama-8B & \textbf{85.37} & 83.62 & 83.62 & 83.62 & S1 \\
GSM8K & Qwen3-4B-2507 & \textbf{10.16} & 3.87 & 9.63 & 9.63 & gate \\
Corr2Cause & Qwen2.5-Coder-7B & \textbf{81.66} & 40.43 & 80.50 & 80.50 & gate \\
BoolQ & DS-R1-Llama-70B & \textbf{87.65} & 87.50 & 87.20 & 87.20 & gate \\
\bottomrule
\end{tabular}
\caption{Representative outcomes generated from the complete summary panel. Bold marks the largest recorded accuracy in each row, not a statistically established improvement. DS-R1 abbreviates DeepSeek-R1-Distill.}
\label{tab:policy-selector-detail}\end{table*}

The policy comparison distinguishes revision gains from the additional benefit of selection. On GSM8K, Llama-3.1-8B improves from 17.44\% to 42.99\% with unconditional revision and to 43.59\% with gating. The gate adds 0.60 percentage points beyond revision, while revision supplies most of the improvement over the initial answer. For Qwen2.5-7B on BoolQ, unconditional revision reduces accuracy from 80.83\% to 72.32\%, but the gate reaches 86.02\%. This comparison suggests that the initial-answer features can distinguish useful from harmful revision in some settings.

Selection also misses opportunities. Qwen3-Coder-30B-A3B on Corr2Cause rises from 56.06\% to 83.70\% under unconditional revision, whereas its gate remains at 56.06\%. DeepSeek-R1-Distill-Qwen-7B similarly reaches 77.78\% with revision but only 71.28\% with gating. In both settings, the gate forgoes gains available from unconditional revision. Conversely, Qwen2.5-7B performs best on GSM8K by keeping its initial answer. Small differences, such as 79.70\% gated versus 79.52\% revised for Qwen2-57B on Corr2Cause, should not be read as established superiority without paired uncertainty.

\begin{figure}[t]
\centering
\includegraphics[width=\linewidth]{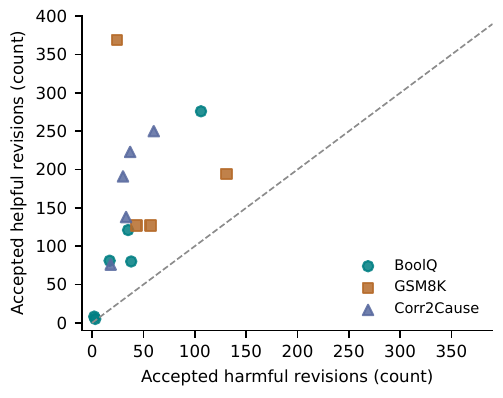}
\caption{Accepted transitions in the 15 available gate-detail summaries. Each equal-sized marker is one model--task setting; shape and color identify the task. Above the diagonal, accepted recoveries exceed accepted harmful flips, implying a gain over \stagezero{} for a matched population. This does not establish a gain over unconditional \stageone{}.}
\label{fig:gate-help-harm}
\end{figure}

\begin{table*}[t]
\centering\small
\begin{tabular}{lrrrrlr}
\toprule
Dataset & Settings & S1$<$S0 & Selected $\geq$ S0 & Percentage & S0/S1/gate & Median saved\\
\midrule
\csname @@input\endcsname generated/aggregate_rows.tex
\bottomrule
\end{tabular}
\caption{Aggregates recomputed from the same 82 rows as Appendix~\ref{app:full-results}. The policy mix counts recorded choices; selected accuracy is not the maximum of the three evaluation accuracies. Savings are percentages of additional \stageone{} generated tokens, with medians computed from the source's whole-percentage values.}
\label{tab:gate-summary}
\end{table*}

BoolQ uses the gate in 24 of 28 recorded policy choices, while GSM8K splits between six initial-answer policies, eleven unconditional revisions, and eleven gates. Corr2Cause uses unconditional revision in eight of 26 settings. Selected accuracy falls below the initial baseline in three BoolQ, three GSM8K, and four Corr2Cause settings. These residual failures show that setting-level policy selection does not guarantee improvement on evaluation examples.

\subsection{Harm and inference-cost preferences}
The reported comparisons weight a recovery and a harmful flip equally. More generally, if $r_1(x)$ is normalized additional revision cost, the utility relative to keeping the initial answer is
\[
\begin{aligned}
\Delta U(g;\lambda,\eta)
&=P(g=1,\wc)\\
&\quad-\lambda P(g=1,\cw)\\
&\quad-\eta\,\mathbb{E}[g(x)r_1(x)].
\end{aligned}
\]
Here $\lambda$ prices harm relative to recovery and $\eta$ prices inference. The experiments evaluate only the accuracy-aligned case $\lambda=1,\eta=0$. Reported token savings are replay estimates relative to generating every revision:
\[
\mathrm{SavedTokens}
=1-\frac{\sum_x g(x)t_1(x)}{\sum_x t_1(x)},
\]
where $t_1(x)$ is the generated-token count of the recorded revision; this excludes the initial response, gate overhead, and prompt-prefill cost. A harm--cost sweep would address different deployment preferences; it is not part of the reported experiment.

\section{Related Work}
\label{sec:related-work}
\paragraph{Outcome analyses of intrinsic correction.}
\citet{huang2023large} document failures of reasoning self-correction without external feedback. \citet{kamoi2024can} distinguish evaluation settings that are often grouped under self-correction, and \citet{zhang2024understanding} examine prompt bias and unstable revisions. \citet{tyen2024mistake} separate locating a mistake from correcting a known one. \citet{yang2025confidence} enumerate initial and revised correctness states and decompose correction into confidence and critique, distinguishing the preservation of correct answers from the correction of errors. We use that accounting to ask whether pre-revision signals can support invocation decisions and whether a learned gate improves on both unconditional policies. Task-sensitive results from \citet{stav2026when} further motivate evaluating a specified model--task--protocol combination.

\paragraph{Feedback, verification, and search.}
Self-Refine iterates feedback and revision~\citep{madaan2023selfrefine}, while Reflexion reuses verbal feedback across trials~\citep{shinn2023reflexion}. CRITIC adds tool-based checks~\citep{gou2024critic}, and Chain-of-Verification generates and answers verification questions~\citep{dhuliawala2023chain}. \citet{weng2023selfverification} rank candidates using backward verification; \citet{xie2023selfevaluation} guide search using stepwise self-evaluation. SelfCheckGPT compares stochastic samples to detect hallucination~\citep{manakul2023selfcheckgpt}. These methods allocate evidence or computation to checking and selection. Our invocation gate uses a narrower information set because it acts before the fixed second generation.

\paragraph{Learned correction and within-generation reflection.}
A trained corrector can improve an imperfect generator~\citep{welleck2023selfcorrect}. SCoRe learns multi-turn correction through reinforcement learning~\citep{kumar2025score}, while S$^2$R trains self-verification and correction with outcome- and process-level signals~\citep{ma2025s2r}. Self-reflective generation intervenes at uncertain token positions using a corrective vector fitted to the generated context~\citep{mu2026selfreflective}. These approaches improve the correction operation or generation process. We hold that operation fixed and examine its invocation. Including reasoning-tuned checkpoints in the panel tests additional explicit revision after their existing deliberation; it does not isolate a causal effect of reasoning training.

\section{Scope and Limitations}
\label{sec:limitations}
The model panel covers open-weight checkpoints up to 123B parameters and three benchmarks. It excludes commercial frontier models, larger open-weight systems, and a harder modern benchmark. Coverage is incomplete, and the prompt experiment is restricted to BoolQ. Neither the task-level comparisons nor the reasoning-model subset establish cross-domain transfer or a causal effect of model training.

Result provenance limits the strength of the policy claims. The recovered paired records validate two GSM8K settings, not the full panel. The version-2 trainer supports training-set cross-validation followed by separate evaluation, but not all historical gate and selector summaries have been linked to frozen configurations and disjoint IDs. These summaries support descriptive comparisons, but the unresolved provenance prevents treating selector gains as verified unbiased estimates. Inconsistent coverage fields are withheld pending matched-record reconstruction.

Protocol choices also affect interpretation. Invalid answers, KEEP/CHANGE application, answer probing, and budget-limited traces can change measured transitions. Count-based intervals condition on a fixed run and parser; they do not include selection or generation uncertainty, and prompt-level intervals are unavailable. Token savings measure avoided revision output tokens; we did not evaluate end-to-end latency or energy. Finally, fitting and selecting a gate requires labeled development data. Transfer to new models or tasks, sensitivity to that calibration burden, and explicit harm--cost preferences remain to be evaluated.

\section{Conclusion}
\label{sec:discussion-conclusion}

Intrinsic self-correction is not simply a question of whether a second answer is more accurate than the first. Across the settings we evaluate, revision both repairs errors and overturns correct answers, and the balance between these outcomes varies by model, task, and refinement prompt. Aggregate accuracy therefore gives only a partial view of what revision changes.
This transition view also changes how we should evaluate selective self-correction. In some settings, an instance-level gate preserves useful corrections while avoiding harmful revisions; in others, always keeping the initial answer or always accepting the revision is the better policy. The value of selective ISC therefore lies not in revision itself, but in deciding when additional inference is worth using. Evaluating that decision requires accounting for both recovered and introduced errors under the exact revision protocol that will be deployed.

\section*{Ethics Statement}

This work evaluates existing open-weight language models on public benchmarks and does not involve human subjects, personal data, or creating new datasets. The selective gating policies studied here are intended to reduce harmful answer changes during inference; they do not introduce new model capabilities or enable adversarial applications. All evaluated models are publicly released under their respective licenses. ISC behavior may differ in deployment contexts not covered by our benchmarks, and practitioners should validate transition rates under their specific use cases before relying on the policies studied here.

\bibliography{ref}

\appendix
\section{Reproducibility Details}
\label{app:reproducibility}
\begin{table}[h]
\centering\small
\begin{tabular}{llrl}
\toprule
Dataset & Evaluation & Size & Development\\
\midrule
BoolQ & validation & 3,270 & train: 9,427\\
GSM8K & test & 1,319 & train: 7,473\\
Corr2Cause & test & 2,246 & dev: 2,246\\
\bottomrule
\end{tabular}
\caption{Nominal dataset splits. These sizes do not certify the denominator of a summary-only gate result.}
\label{tab:dataset-splits}
\end{table}

The retained panel includes Llama, Qwen, Gemma, Yi, OLMo, Mixtral, Devstral, GPT-OSS, DeepSeek, and Phi checkpoints, including code-oriented and reasoning-tuned variants. The complete tables name every retained checkpoint. Smaller checkpoints and StarCoder2 appear only in the separate diagnostic panels. We do not pool those panels with the 82-row comparison.

\subsection{Compute environment}
We submitted most generation jobs to a remote Ray cluster. Large-model runs that required model sharding used a single four-GPU Ray node with NVIDIA A100 80GB GPUs. These jobs used \texttt{bfloat16} inference, Hugging Face \texttt{device\_map=auto} placement, and an effective loader cap of about 72~GiB per GPU, with CPU offload caps of 160-256~GiB when needed. Smaller models used one- or two-GPU Ray actors under the same generation and parsing code. We trained gate models separately on CPU; we report the GPU node specification only to document the inference testbed, and it should not be interpreted as a throughput benchmark.

The inference pipeline uses PyTorch and the Hugging Face \texttt{transformers}/\texttt{accelerate} packages. Model weights and tokenizers are loaded from Hugging Face model repositories using the corresponding tokenizer and causal-language-model loaders; generation uses the standard autoregressive generation interface, and likelihood or confidence features are computed from recorded logits or scoring passes under the same serialized prompt. Dataset splits are loaded through Hugging Face \texttt{datasets} when available and then materialized as JSONL records containing raw generations, parsed answers, correctness labels, and generation metadata. Each run's metadata file records exact package versions, model identifiers, tokenizer names, prompt settings, and memory caps.

\section{Protocol and Parsing Audits}
\label{app:protocol}
The paired-record audit counts every matched item, including invalid answers. Both recovered GSM8K pairs have unique IDs and consistent initial predictions and correctness labels across stage files. The source inventory records file hashes, and the audit script recomputes all counts in Table~\ref{tab:protocol-checks}. Matching the summary's rounded accuracies corroborates these initial/revised outcomes; it does not establish provenance for a gate fitted on the same checkpoint.

\paragraph{Refinement prompt templates.}
The \stageone{} prompt wraps the original question, the previous final answer, and a refinement instruction. The default neutral instruction is: ``Let's quickly double-check the reasoning above. Don't change the answer unless contradicted by the question.'' For arithmetic tasks, the conservative verifier instruction is: ``Act as a verifier, not a fresh solver. The previous final answer is the default. Keep it unless you can identify a concrete arithmetic or constraint contradiction. If the previous answer is valid, output exactly: \texttt{KEEP FINAL: <previous answer>}. Only if it is wrong, output exactly: \texttt{CHANGE FINAL: <corrected answer>}. Do not output any other text.'' For Corr2Cause, the task-adapted instruction is: ``Reconsider whether the correlational statements logically support the hypothesis. Do not rely on world knowledge. If the premise does not justify the prior answer, correct it. Output only \texttt{FINAL: YES} or \texttt{FINAL: NO}.'' For answer-probe reasoning models, the probe prompt asks: ``Please verify the previous answer step by step. If it is correct, keep it. If it is wrong, correct it.''

The prompt-sensitivity study uses five controlled refinement families. The neutral prompt double-checks internal consistency and instructs the model not to change unless the passage clearly contradicts the initial answer. The stay and hard-stay prompts increasingly encourage retaining the initial answer unless there is direct or explicit contradiction. The switch and hard-switch prompts increasingly encourage moving away from the initial answer when the previous reasoning has any plausible weakness. All variants preserve the same final-answer contract, requiring exactly one yes/no final answer.

We use several controlled prompt variants when the default contract is unreliable or when the experiment explicitly studies prompt sensitivity. Strict-retry variants append a short instruction requiring exactly one final-answer line and are used only to repair missing or malformed final answers. Chat-template variants serialize the same logical prompt using the tokenizer's chat template; we then compute likelihoods, margins, entropy features, and gate inputs under the same serialized prefix. Verifier-style variants, used mainly for GSM8K diagnostics, ask the model to either keep the previous answer or emit a corrected candidate; the appendix reports both the applied KEEP/CHANGE rule and the direct candidate score when they differ. For reasoning-tuned models whose native traces are long or whose final answer is not emitted in the expected format, answer-probe variants preserve the generated trace and then query for a compact final answer under a fixed answer prefix.

The parser is deliberately task-specific and conservative. For BoolQ and Corr2Cause, it maps explicit yes/no final answers to binary labels. For GSM8K, it extracts the final integer answer after normalization of commas, signs, and surrounding text. Outputs with no valid final answer are marked as parse failures and count as incorrect in the recovered paired records. These parse states are retained as metadata and, where available before \stageone{}, may be used as gate features; they are also reported as protocol diagnostics to avoid conflating formatting artifacts with failures of semantic self-correction.

\paragraph{Serialisation and trace completion.}
Generation and candidate scoring must use the same serialized prefix to make their confidence features comparable. The recovered ordinary OLMo run uses a raw prompt, whereas the Phi reasoning run uses its tokenizer's chat template. For trace-plus-probe generation, a compact answer after a budget-limited trace remains a scored output, but does not establish that the underlying reasoning completed. The main-text counts report these conditions explicitly.

\paragraph{Answer-selection sensitivity.}
The separate verifier diagnostics in Table~\ref{tab:policy-sensitivity} compare applied KEEP/CHANGE answers with extracted candidates. They use their own parsing conditions and must not be substituted for the retained accuracy-panel values. In particular, equal initial and applied accuracy can reflect an answer-selection rule that preserves the initial answer even when the generated candidate changes.

\begin{table*}[h]
\centering
\scriptsize
\begin{tabular}{lrrrrrr}
\toprule
Model & Acc0 & Applied acc. & Candidate acc. & KEEP count & CHANGE count & Failed-change count \\
\midrule
Llama-3.1-8B-Instruct & 17.5 & 43.0 & 44.8 & 568 & 731 & 20 \\
Qwen2.5-32B-Instruct & 37.9 & 37.1 & 0.0 & 586 & 69 & 664 \\
Olmo-3.1-32B-Instruct & 47.5 & 48.6 & 26.8 & 1225 & 26 & 68 \\
\bottomrule
\end{tabular}
\caption{Answer-selection sensitivity on representative GSM8K verifier-style runs. ``Applied acc.'' follows the answer-selection rule, whereas ``Candidate acc.'' scores the parsed \stageone{} candidate directly when present. This is a protocol diagnostic, not evidence that parser or action-format artifacts are semantic ISC failures. Denominators may differ slightly across columns because each rule has its own parseability condition.}
\label{tab:policy-sensitivity}
\end{table*}
\section{Gate Features and Implementation}
\label{app:diagnostics}
\label{app:gate-details}
The recovered version-2 implementation builds features from the initial generation, removes unusable columns using training data, and obtains out-of-fold predictions for threshold selection. Its separate help/harm models use correctness transitions as targets. It then refits on training data and evaluates frozen thresholds on a supplied evaluation set. In that code path, it uses evaluation labels to calculate outcome metrics. The experiment runner must check the ID-overlap report; the script does not enforce disjointness.

The code supports logistic regression, HistGradientBoosting, and LightGBM. The retained method description specifies a 16-trial search, starting with defaults and then sampling configurations. Logistic regression uses median imputation, standardization, class-balanced weights, the \texttt{lbfgs} solver, 2,000 maximum iterations, and $C\in[10^{-2},10^2]$. HistGradientBoosting samples 150--500 iterations, depth 3--10, learning rate $10^{-2}$--$10^{-0.5}$, and $L_2$ regularisation $10^{-6}$--$10^1$. LightGBM samples 200--800 estimators, learning rate $10^{-2}$--$10^{-0.5}$, 15--95 leaves, and subsample and column-sample rates in $[0.6,1.0]$. Defaults are $C=1$ for logistic regression; 300 iterations and learning rate 0.1 for HistGradientBoosting; and 400 estimators, learning rate 0.05, 31 leaves, and 0.8 sampling rates for LightGBM. A saved configuration is needed to identify which path and parameters produced a historical row.

\begin{table*}[h]
\centering
\scriptsize
\resizebox{\textwidth}{!}{%
\begin{tabular}{@{}lll@{}}
\toprule
Feature group & Examples & Role in the learned gate \\
\midrule
\begin{tabular}[t]{@{}l@{}}Parser and final-answer\\format\end{tabular} &
\begin{tabular}[t]{@{}l@{}}parse OK, final present,\\final count, final-line length\end{tabular} &
\begin{tabular}[t]{@{}l@{}}Captures whether the initial answer satisfies the task contract and\\whether refinement may mainly repair formatting or extraction risk.\end{tabular} \\
\begin{tabular}[t]{@{}l@{}}Prompt and generation\\metadata\end{tabular} &
\begin{tabular}[t]{@{}l@{}}prompt length, prompt truncation,\\generation length, post-final truncation\end{tabular} &
\begin{tabular}[t]{@{}l@{}}Records length and truncation conditions visible after \stagezero{}\\that may signal unreliable or incomplete initial outputs.\end{tabular} \\
\begin{tabular}[t]{@{}l@{}}Generation log-probability\\summaries\end{tabular} &
\begin{tabular}[t]{@{}l@{}}mean/min/last log-probability,\\log-probability standard deviation\end{tabular} &
\begin{tabular}[t]{@{}l@{}}Measures confidence and instability in the emitted \stagezero{} text\\when token log-probabilities are available.\end{tabular} \\
\begin{tabular}[t]{@{}l@{}}Binary-answer confidence\\features\end{tabular} &
\begin{tabular}[t]{@{}l@{}}yes/no confidence, entropy,\\logit margin, probability gap\end{tabular} &
\begin{tabular}[t]{@{}l@{}}Used for BoolQ and Corr2Cause to estimate whether the initial\\yes/no answer is uncertain enough to justify refinement.\end{tabular} \\
\begin{tabular}[t]{@{}l@{}}Integer-answer candidate\\features\end{tabular} &
\begin{tabular}[t]{@{}l@{}}integer confidence, entropy,\\logit margin, number of candidates\end{tabular} &
\begin{tabular}[t]{@{}l@{}}Used for GSM8K when candidate scores are available to capture\\uncertainty over numeric final answers.\end{tabular} \\
\begin{tabular}[t]{@{}l@{}}Parsed-answer consistency\\features\end{tabular} &
\begin{tabular}[t]{@{}l@{}}parsed/scored answer match,\\signed margin, parsed integer is best\end{tabular} &
\begin{tabular}[t]{@{}l@{}}Checks whether the parsed answer agrees with the model's scored\\candidate preference under the same \stagezero{} prompt.\end{tabular} \\
\bottomrule
\end{tabular}
}
\caption{Feature groups used by the reported learned gates. Examples are shortened aliases for the corresponding implementation fields. All features are pre-\stageone{} signals extracted from the initial generation or its scoring pass. The reported gates do not use post-\stageone {} agreement, refined-answer confidence, trace completion, or gold-label diagnostics.}
\label{tab:feature-availability}
\end{table*}
\subsection{Numerical consistency checks}
For every complete-panel row, the audit verifies that selected accuracy equals the recorded selected policy's accuracy. It does not choose the maximum evaluation accuracy after the fact. For each of the 15 accepted-transition rows, it also tests the gate identity at the nominal task denominator and records that denominator linkage remains unverified. All 15 count gains are compatible with the one-decimal accuracies under that assumption; this arithmetic agreement is not proof of shared item IDs.

Three reported 100.0\%-coverage rows fail the endpoint identity even allowing for one-decimal accuracy and coverage rounding: DeepSeek-R1-Distill-Qwen-14B and -7B on GSM8K, and Phi-4-reasoning-plus on Corr2Cause. DeepSeek-R1-Distill-Llama-8B on GSM8K also differs from \stageone{} at the displayed precision, although the rounded values alone do not exclude a near-full-coverage explanation. The displayed table omits coverage pending verification of these values. For the Qwen-14B transition summary, comparing 130 total recoveries and 44 total harms with 127 and 43 accepted ones would imply at least four blocked examples if the rows were item-matched. The available summaries do not establish that match.

The source files preserve both the complete-panel token-savings column and the different savings reported by the smaller gate-detail summary. These are not merged. All main aggregates use the complete-panel column, whose values are rounded to whole percentages. A reported tie is retained as a tie at that precision; the policy field preserves the recorded selection without imposing a new tie-break rule. Missing gate values denote unavailable or failed gate results and retain the recorded unconditional fallback.

\begin{table*}[t]
\centering\small
\setlength{\tabcolsep}{3pt}
\begin{tabular}{llrrr}
\toprule
Dataset & Model & Gated accuracy & Accepted $\wc$ & Accepted $\cw$\\
\midrule
BoolQ & Qwen2.5-7B & 86.0 & 276 & 106 \\
BoolQ & gpt-oss-20b & 85.1 & 121 & 35 \\
BoolQ & Olmo-3.1-32B & 87.8 & 81 & 17 \\
BoolQ & Phi-4-mini-reasoning & 83.6 & 80 & 38 \\
BoolQ & Phi-4-reasoning-plus & 88.2 & 8 & 2 \\
BoolQ & Qwen2.5-72B & 89.0 & 5 & 3 \\
GSM8K & Llama-3.1-8B & 43.6 & 369 & 24 \\
GSM8K & DS-R1-Qwen-14B & 85.9 & 127 & 43 \\
GSM8K & DS-R1-Qwen-7B & 77.6 & 127 & 57 \\
GSM8K & DS-R1-Llama-8B & 59.4 & 194 & 131 \\
Corr2Cause & Olmo-3-7B & 80.9 & 223 & 37 \\
Corr2Cause & Phi-4-reasoning-plus & 71.3 & 191 & 30 \\
Corr2Cause & Qwen2-57B-A14B & 79.7 & 250 & 60 \\
Corr2Cause & DS-R1-Qwen-7B & 71.3 & 138 & 33 \\
Corr2Cause & Qwen2.5-32B & 82.5 & 76 & 18 \\
\bottomrule
\end{tabular}
\caption{Separate accepted-transition summaries underlying Figure~\ref{fig:gate-help-harm}. Gated accuracy is a percentage. Coverage is withheld because the supplied summary contains inconsistent endpoints; source values and automatic checks are retained in the accompanying CSV files.}
\label{tab:gate-results}
\end{table*}
\begin{table}[h]
\centering
\scriptsize
\setlength{\tabcolsep}{3pt}
\renewcommand{\arraystretch}{0.92}
\begin{tabular}{llrrrrr}
\toprule
Dataset & Feature group & Runs & $>$S0 & $>$both & Gain & Saved \\
\midrule
BoolQ & conf/logprob & 17 & 12 & 10 & +0.4 & 69.2 \\
BoolQ & parser/format & 23 & 19 & 15 & +0.4 & 54.8 \\
BoolQ & trace/protocol & 13 & 9 & 4 & +0.1 & 76.4 \\
BoolQ & full gate & 25 & 21 & 19 & +0.5 & 57.2 \\
GSM8K & conf/logprob & 14 & 9 & 4 & +0.1 & 50.5 \\
GSM8K & parser/format & 18 & 14 & 2 & +0.6 & 0.1 \\
GSM8K & trace/protocol & 9 & 6 & 2 & +0.1 & 0.0 \\
GSM8K & full gate & 28 & 15 & 5 & +0.1 & 7.6 \\
Corr2C. & conf/logprob & 15 & 7 & 1 & +0.0 & 24.7 \\
Corr2C. & parser/format & 19 & 15 & 5 & +1.1 & 0.0 \\
Corr2C. & trace/protocol & 13 & 8 & 1 & +0.3 & 0.0 \\
Corr2C. & full gate & 25 & 15 & 5 & +0.3 & 3.4 \\
\bottomrule
\end{tabular}
\caption{Feature-restricted baseline comparisons. Panels differ in coverage and predictor class, so these are not controlled ablations. Gain is the median gated accuracy improvement over \stagezero{} on the reported evaluation data, in percentage points. Saved is the median replay estimate of generated \stageone{} tokens avoided, relative to always using refinement. Feature-only rows use simple one-dimensional thresholds selected from the training split, while ``full gate'' reports the selected tabular gate suite.}
\label{tab:gate-ablation}
\end{table}
\section{Bootstrap Uncertainty}
\label{app:bootstrap}
The transition intervals below are multinomial-bootstrap intervals from the reported four-state counts. They condition on the chosen run and extraction rule and omit uncertainty from model sampling, prompt choice, and run selection. The main text includes the intervals for its two GSM8K examples. These count-based intervals do not certify a summary's provenance or resolve incompatible gate-coverage values.

\begin{table*}[h]
\centering
\scriptsize
\begin{tabular}{llrr}
\toprule
Model & Task & Net benefit 95\% CI & Harm rate 95\% CI \\
\midrule
Olmo-3.1-32B-Instruct & BoolQ & [+1.0, +2.4] & [1.0, 1.9] \\
Llama-3.1-70B-Instruct & BoolQ & [-2.1, -0.1] & [4.6, 6.3] \\
Llama-3.1-8B-Instruct & GSM8K & [+22.9, +28.4] & [14.2, 24.3] \\
DeepSeek-R1-Distill-Qwen-14B & GSM8K & [+4.6, +8.4] & [3.0, 5.5] \\
DeepSeek-R1-Distill-Qwen-7B & Corr2Cause & [+9.6, +12.7] & [2.5, 4.3] \\
Phi-4-reasoning-plus & Corr2Cause & [+5.9, +8.7] & [2.5, 4.4] \\
\bottomrule
\end{tabular}
\caption{Bootstrap intervals for the representative transition rows. Net benefit and harm rates are expressed in percentage points. Intervals are resampled from observed transition counts and do not include run-selection uncertainty.}
\label{tab:transition-bootstrap}
\end{table*}
\section{Separate Diagnostic Panels}
\label{app:diagnostic-panels}
We retain these three original two-stage plots as protocol diagnostics. Their source CSVs mix summary populations, including combined training/evaluation summaries and evaluation-only runs; they also contain checkpoints outside the retained panel. The GSM8K Llama-3.1-70B diagnostic shows roughly 35.5\% to 8.3\%, whereas the retained summary reports 35.48\% to 36.09\%. Their scoring/run linkage is not established, so the diagnostic drop does not support the retained setting. Figure~\ref{fig:two-stage-isc} is the source-consistent comparison used in the main text.

\begin{figure*}[p]
\centering
\includegraphics[width=\textwidth]{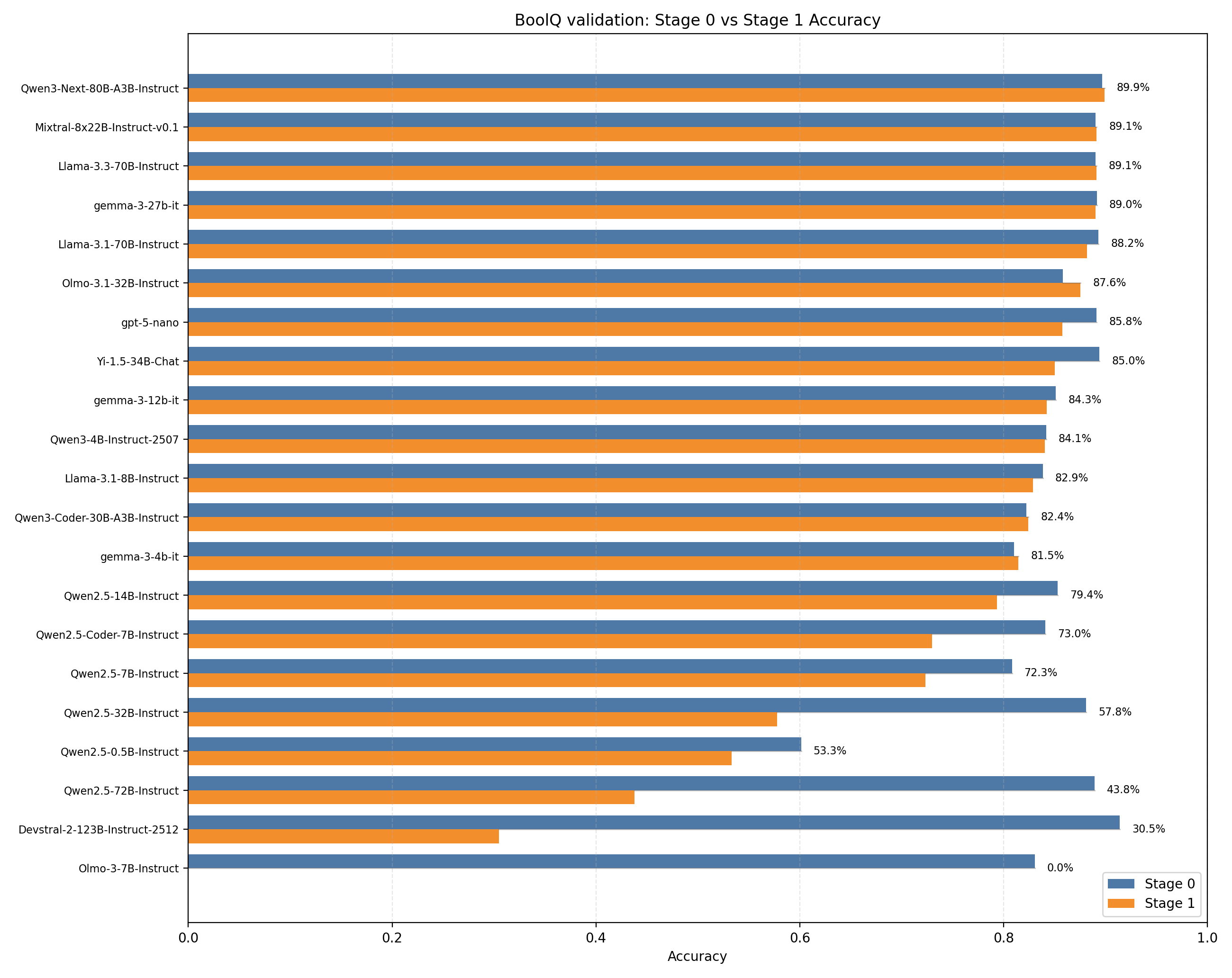}
\caption{Original BoolQ diagnostic panel, using its separate summary populations. This is not the 28-setting retained policy panel.}
\end{figure*}
\begin{figure*}[p]
\centering
\includegraphics[width=\textwidth]{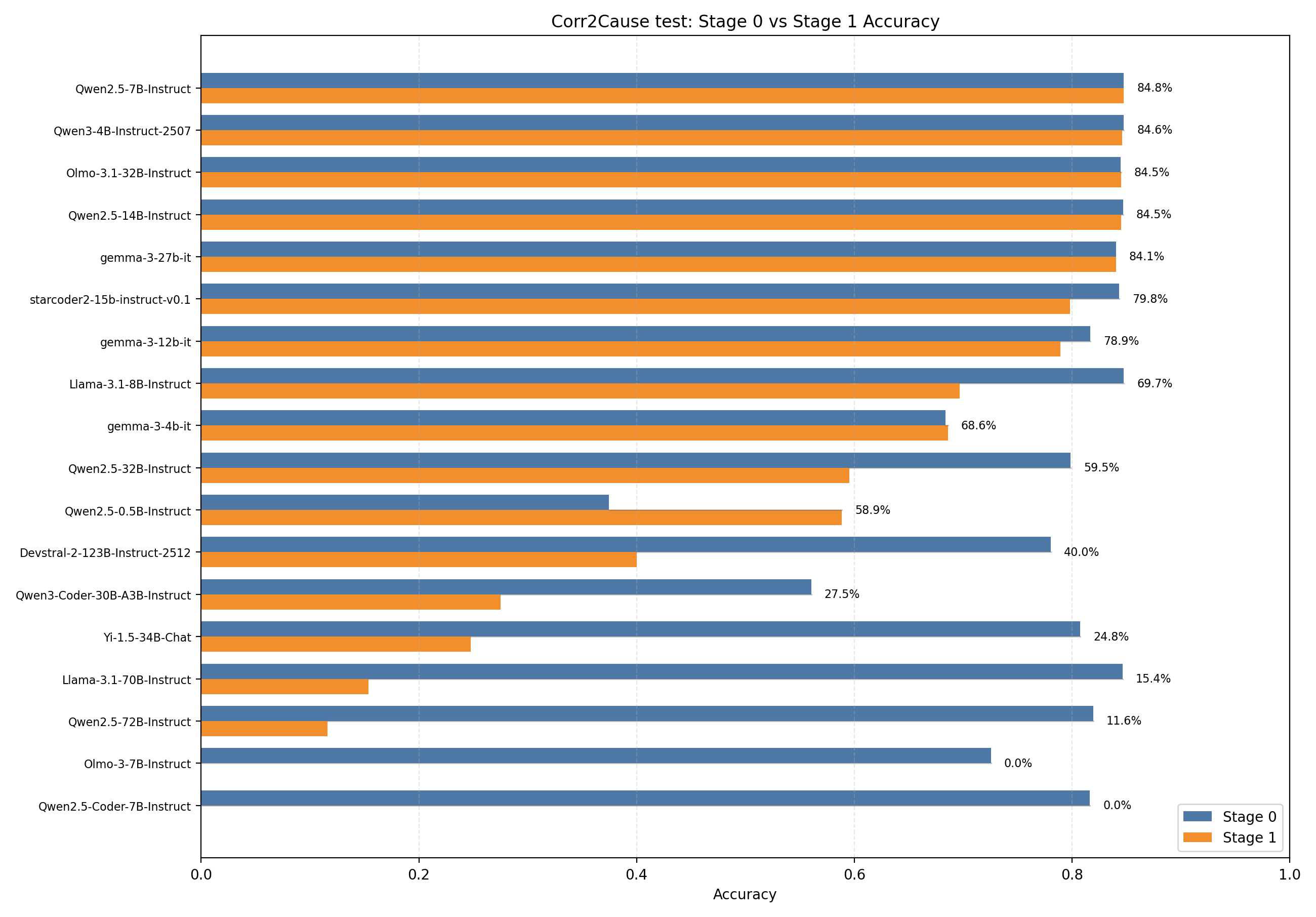}
\caption{Original Corr2Cause diagnostic panel. Run populations and model coverage differ from the retained results.}
\end{figure*}
\begin{figure*}[p]
\centering
\includegraphics[width=\textwidth]{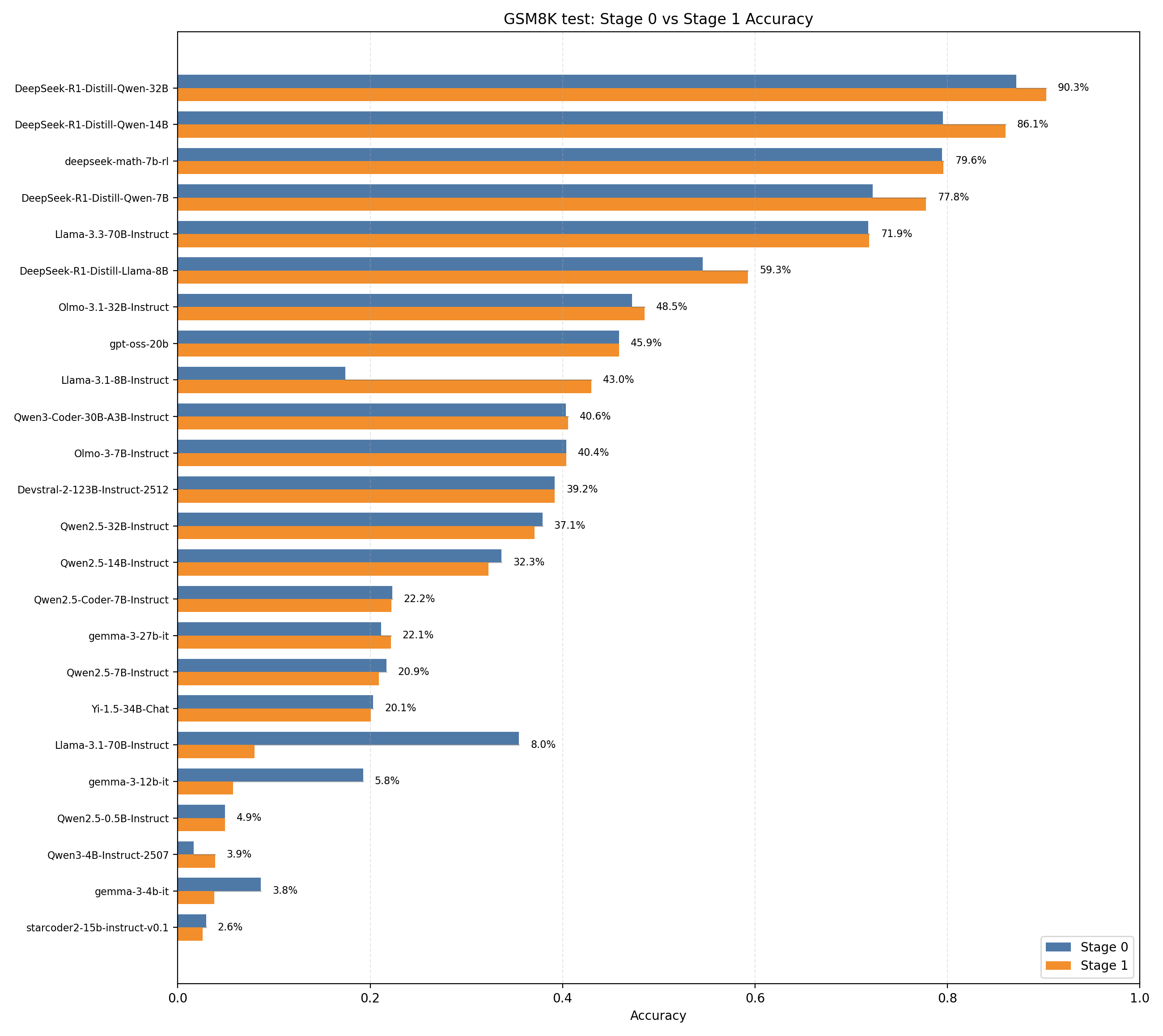}
\caption{Original GSM8K diagnostic panel. Its large Llama-3.1-70B drop must not be equated with the retained run's positive change.}
\end{figure*}

\clearpage
\onecolumn
\section{Complete Per-Model Policy Results}
\label{app:full-results}
\label{tab:full-policy-selector-results}
These 82 rows are the source for the main accuracy figure and aggregate policy table. They transcribe the complete policy summary; they are not newly run experiments. Initial, revised, gated, and selected accuracies are percentages. $N$ is left as a dash because item-level gate denominators have not been linked to these rows; nominal benchmark sizes must not be silently substituted. A dash in Gate indicates an unavailable or failed gate result. S0 and S1 denote unconditional initial and revised policies. Selected accuracy follows the recorded policy even when another policy has higher evaluation accuracy. Saved is the percentage of additional revision output tokens avoided by that selected policy, at the source's whole-percentage precision. DS-R1 abbreviates DeepSeek-R1-Distill; instruction suffixes are shortened.

\begingroup
\fontsize{9}{11}\selectfont
\setlength{\tabcolsep}{4pt}
\begin{table}[ht]\centering\fontsize{9.5}{11.5}\selectfont
\caption{BoolQ: complete summary panel. Accuracies and savings are percentages.}\label{tab:full-boolq}
\begin{tabular}{@{}lrrrrlrr@{}}
\toprule Model & $N$ & Initial & Revised & Gate & \shortstack{Selected\\policy} & \shortstack{Selected\\accuracy} & \shortstack{Saved tokens\\(\%)}\\
\midrule
DS-R1-Llama-70B & --- & 87.65 & 87.50 & 87.20 & gate & 87.20 & 20 \\
DS-R1-Llama-8B & --- & 85.37 & 83.62 & 83.62 & S1 & 83.62 & 0 \\
DS-R1-Qwen-14B & --- & 87.16 & 87.09 & 88.13 & gate & 88.13 & 56 \\
DS-R1-Qwen-32B & --- & 88.90 & 88.07 & 89.33 & gate & 89.33 & 57 \\
DS-R1-Qwen-7B & --- & 77.77 & 74.80 & 78.81 & gate & 78.81 & 52 \\
Devstral-2-123B-2512 & --- & 91.41 & 30.49 & --- & S0 & 91.41 & 100 \\
Llama-3.1-70B & --- & 89.33 & 88.20 & 90.73 & gate & 90.73 & 56 \\
Llama-3.1-8B & --- & 83.85 & 82.87 & 85.02 & gate & 85.02 & 86 \\
Llama-3.3-70B & --- & 89.02 & 89.11 & 89.02 & gate & 89.02 & 92 \\
Mixtral-8x22B-v0.1 & --- & 89.05 & 89.14 & --- & S1 & 89.14 & 0 \\
Olmo-3.1-32B & --- & 85.84 & 87.55 & 87.80 & gate & 87.80 & 55 \\
Olmo-3-7B & --- & 83.06 & 82.32 & 83.58 & gate & 83.58 & 78 \\
Phi-4-mini-reasoning & --- & 82.32 & 81.77 & 83.61 & gate & 83.61 & 59 \\
Phi-4-reasoning & --- & 84.05 & 74.54 & 85.89 & gate & 85.89 & 52 \\
Phi-4-reasoning-plus & --- & 88.01 & 87.43 & 88.20 & gate & 88.20 & 56 \\
Qwen2.5-14B & --- & 85.32 & 79.36 & 86.64 & gate & 86.64 & 53 \\
Qwen2.5-32B & --- & 88.10 & 87.40 & 88.26 & gate & 88.26 & 86 \\
Qwen2.5-72B & --- & 88.93 & 89.08 & 88.99 & gate & 88.99 & 96 \\
Qwen2.5-7B & --- & 80.83 & 72.32 & 86.02 & gate & 86.02 & 71 \\
Qwen2.5-Coder-7B & --- & 84.10 & 84.98 & 84.83 & gate & 84.83 & 6 \\
Qwen2-57B-A14B & --- & 89.25 & 89.22 & 89.28 & gate & 89.28 & 98 \\
Qwen3-4B-2507 & --- & 84.19 & 84.07 & 85.08 & gate & 85.08 & 45 \\
Qwen3-Coder-30B-A3B & --- & 82.23 & 82.45 & --- & S1 & 82.45 & 0 \\
Qwen3-Next-80B-A3B & --- & 89.66 & 89.91 & 89.97 & gate & 89.97 & 22 \\
Yi-1.5-34B-Chat & --- & 89.42 & 85.02 & 89.48 & gate & 89.48 & 94 \\
gemma-3-12b-it & --- & 85.14 & 84.25 & 85.26 & gate & 85.26 & 80 \\
gemma-3-27b-it & --- & 89.17 & 89.02 & 89.14 & gate & 89.14 & 96 \\
gpt-oss-20b & --- & 82.45 & 84.13 & 85.08 & gate & 85.08 & 80 \\
\bottomrule\end{tabular}\end{table}

\clearpage
\begin{table}[ht]\centering\fontsize{9.5}{11.5}\selectfont
\caption{GSM8K: complete summary panel. Accuracies and savings are percentages.}\label{tab:full-gsm8k}
\begin{tabular}{@{}lrrrrlrr@{}}
\toprule Model & $N$ & Initial & Revised & Gate & \shortstack{Selected\\policy} & \shortstack{Selected\\accuracy} & \shortstack{Saved tokens\\(\%)}\\
\midrule
DS-R1-Llama-70B & --- & 89.20 & 90.80 & 90.80 & S1 & 90.80 & 0 \\
DS-R1-Llama-8B & --- & 54.59 & 59.29 & 59.36 & S1 & 59.29 & 0 \\
DS-R1-Qwen-14B & --- & 79.53 & 86.05 & 85.90 & S1 & 86.05 & 0 \\
DS-R1-Qwen-32B & --- & 87.19 & 90.30 & 89.84 & S1 & 90.30 & 0 \\
DS-R1-Qwen-7B & --- & 72.25 & 77.79 & 77.56 & S1 & 77.79 & 0 \\
Devstral-2-123B-2512 & --- & 39.20 & 39.20 & 39.20 & gate & 39.20 & 88 \\
Llama-3.1-70B & --- & 35.48 & 36.09 & 36.09 & S1 & 36.09 & 0 \\
Llama-3.1-8B & --- & 17.44 & 42.99 & 43.59 & gate & 43.59 & 9 \\
Llama-3.3-70B & --- & 71.80 & 71.87 & 71.80 & gate & 71.80 & 39 \\
Mixtral-8x22B-v0.1 & --- & 25.08 & 24.62 & 24.92 & S0 & 25.08 & 100 \\
Olmo-3.1-32B & --- & 47.23 & 48.52 & 48.45 & S1 & 48.52 & 0 \\
Olmo-3-7B & --- & 40.41 & 40.41 & 40.41 & S1 & 40.41 & 0 \\
Phi-4-mini-reasoning & --- & 87.26 & 88.63 & 88.63 & gate & 88.63 & 21 \\
Phi-4-reasoning & --- & 95.45 & 95.91 & 95.98 & gate & 95.98 & 6 \\
Phi-4-reasoning-plus & --- & 92.49 & 92.19 & 92.49 & S0 & 92.49 & 100 \\
Qwen2.5-14B & --- & 33.66 & 32.30 & 33.74 & gate & 33.74 & 84 \\
Qwen2.5-32B & --- & 37.91 & 37.07 & 37.98 & gate & 37.98 & 88 \\
Qwen2.5-72B & --- & 40.86 & 40.94 & 40.86 & S0 & 40.86 & 100 \\
Qwen2.5-7B & --- & 21.68 & 20.92 & 20.92 & S0 & 21.68 & 100 \\
Qwen2.5-Coder-7B & --- & 22.29 & 22.21 & 22.21 & S0 & 22.29 & 100 \\
Qwen2-57B-A14B & --- & 24.87 & 23.88 & 24.79 & S0 & 24.87 & 100 \\
Qwen3-4B-2507 & --- & 10.16 & 3.87 & 9.63 & gate & 9.63 & 62 \\
Qwen3-Coder-30B-A3B & --- & 40.33 & 40.56 & 40.56 & S1 & 40.56 & 0 \\
Yi-1.5-34B-Chat & --- & 20.32 & 20.09 & 20.09 & S1 & 20.09 & 0 \\
deepseek-math-7b-rl & --- & 79.45 & 79.61 & 79.45 & gate & 79.45 & 100 \\
gemma-3-12b-it & --- & 17.82 & 5.76 & 17.74 & gate & 17.74 & 96 \\
gemma-3-27b-it & --- & 21.15 & 22.14 & 22.14 & gate & 22.14 & 63 \\
gpt-oss-20b & --- & 45.87 & 45.94 & 45.94 & S1 & 45.94 & 0 \\
\bottomrule\end{tabular}\end{table}

\clearpage
\begin{table}[ht]\centering\fontsize{9.5}{11.5}\selectfont
\caption{Corr2Cause: complete summary panel. Accuracies and savings are percentages.}\label{tab:full-corr2cause}
\begin{tabular}{@{}lrrrrlrr@{}}
\toprule Model & $N$ & Initial & Revised & Gate & \shortstack{Selected\\policy} & \shortstack{Selected\\accuracy} & \shortstack{Saved tokens\\(\%)}\\
\midrule
DS-R1-Llama-8B & --- & 82.63 & 83.25 & 83.56 & gate & 83.56 & 98 \\
DS-R1-Qwen-14B & --- & 80.54 & 81.70 & 81.66 & S1 & 81.70 & 0 \\
DS-R1-Qwen-32B & --- & 80.94 & 81.12 & 81.12 & S1 & 81.12 & 0 \\
DS-R1-Qwen-7B & --- & 66.61 & 77.78 & 71.28 & S1 & 77.78 & 0 \\
Devstral-2-123B-2512 & --- & 78.05 & 40.03 & 80.19 & gate & 80.19 & 74 \\
Llama-3.1-70B & --- & 84.64 & 82.72 & 84.64 & gate & 84.64 & 100 \\
Llama-3.1-8B & --- & 84.77 & 84.77 & 84.77 & S0 & 84.77 & 100 \\
Llama-3.3-70B & --- & 85.13 & 84.77 & 84.82 & gate & 84.82 & 1 \\
Mixtral-8x22B-v0.1 & --- & 83.58 & 85.27 & 84.98 & S1 & 85.27 & 0 \\
Olmo-3.1-32B & --- & 84.46 & 84.51 & 84.51 & S1 & 84.51 & 0 \\
Olmo-3-7B & --- & 72.57 & 81.21 & 80.85 & gate & 80.85 & 1 \\
Phi-4-mini-reasoning & --- & 80.81 & 74.71 & 81.83 & gate & 81.83 & 98 \\
Phi-4-reasoning & --- & 74.00 & 78.32 & --- & S1 & 78.32 & 0 \\
Phi-4-reasoning-plus & --- & 64.16 & 71.46 & 71.33 & gate & 71.33 & 0 \\
Qwen2.5-14B & --- & 84.68 & 84.51 & 84.64 & gate & 84.64 & 26 \\
Qwen2.5-32B & --- & 79.88 & 82.59 & 82.46 & gate & 82.46 & 16 \\
Qwen2.5-72B & --- & 81.97 & 82.81 & 82.24 & gate & 82.24 & 99 \\
Qwen2.5-7B & --- & 84.77 & 84.77 & 84.77 & S0 & 84.77 & 100 \\
Qwen2.5-Coder-7B & --- & 81.66 & 40.43 & 80.50 & gate & 80.50 & 95 \\
Qwen2-57B-A14B & --- & 71.24 & 79.52 & 79.70 & gate & 79.70 & 1 \\
Qwen3-4B-2507 & --- & 84.77 & 84.59 & 84.77 & S0 & 84.77 & 100 \\
Qwen3-Coder-30B-A3B & --- & 56.06 & 83.70 & 56.06 & S1 & 83.70 & 0 \\
Yi-1.5-34B-Chat & --- & 80.77 & 84.86 & 83.48 & S1 & 84.86 & 0 \\
gemma-3-12b-it & --- & 81.70 & 78.94 & 81.57 & gate & 81.57 & 3 \\
gemma-3-27b-it & --- & 84.06 & 84.06 & 84.06 & gate & 84.06 & 0 \\
gpt-oss-20b & --- & 54.41 & 53.16 & 54.67 & gate & 54.67 & 99 \\
\bottomrule\end{tabular}\end{table}

\endgroup
\end{document}